\documentclass[10pt,conference]{IEEEtran}
\IEEEoverridecommandlockouts
\usepackage{cite}
\usepackage{amsmath,amssymb,amsfonts}
\usepackage{booktabs}
\usepackage{multirow}
\usepackage{array}
\usepackage{graphicx}
\usepackage{textcomp}
\usepackage{xcolor}
\usepackage[hidelinks]{hyperref}
\usepackage{orcidlink}
\usepackage{cleveref}

\begin{document}

\bstctlcite{IEEEexample:BSTcontrol}

\title{Hybrid Quantum-inspired Kolmogorov–Arnold Networks for Privacy-Aware Federated Biosignal Learning}

\author{
\IEEEauthorblockN{
    Chun-Hua Lin\IEEEauthorrefmark{1}\IEEEauthorrefmark{2}\orcidlink{0009-0002-4383-0453},
    Samuel Yen-Chi Chen\IEEEauthorrefmark{3}\orcidlink{0000-0003-0114-4826},
    Yu-Chao Hsu\IEEEauthorrefmark{2}\IEEEauthorrefmark{4}\orcidlink{0009-0004-7221-3854},
    Kuo-Chung Peng\IEEEauthorrefmark{1}\IEEEauthorrefmark{2}\orcidlink{0009-0001-8342-2481},
    Jiun-Cheng Jiang\IEEEauthorrefmark{1}\IEEEauthorrefmark{2}\IEEEauthorrefmark{7}\orcidlink{0009-0005-1134-4962},
    \\
    Chi-Sheng Chen\IEEEauthorrefmark{5}\orcidlink{0000-0003-0807-0217},
    Tai-Yue Li\IEEEauthorrefmark{2}\orcidlink{0000-0002-1993-1863},
    Nan-Yow Chen\IEEEauthorrefmark{2}\orcidlink{0000-0001-8139-6809},
    En-Jui Kuo\IEEEauthorrefmark{6}\IEEEauthorrefmark{9}\orcidlink{0000-0002-6770-0285},
    Hsi-Sheng Goan\IEEEauthorrefmark{1}\IEEEauthorrefmark{7}\IEEEauthorrefmark{8}\IEEEauthorrefmark{10}\orcidlink{0000-0001-8117-5846}
}

\IEEEauthorblockA{\IEEEauthorrefmark{1} Department of Physics and Center for Theoretical Physics, National Taiwan University, Taipei, Taiwan}
\IEEEauthorblockA{\IEEEauthorrefmark{2} National Center for High-Performance Computing, National Institutes of Applied Research, Hsinchu, Taiwan}
\IEEEauthorblockA{\IEEEauthorrefmark{3} Brookhaven National Laboratory, Upton, NY, USA}
\IEEEauthorblockA{\IEEEauthorrefmark{4} School of Electrical Engineering, Korea Advanced Institute of Science and Technology, Daejeon, Korea}
\IEEEauthorblockA{\IEEEauthorrefmark{5} Beth Israel Deaconess Medical Center, Harvard University, Boston, MA, USA}
\IEEEauthorblockA{\IEEEauthorrefmark{6} Department of Electrophysics, National Yang Ming Chiao Tung University, Hsinchu, Taiwan}
\IEEEauthorblockA{\IEEEauthorrefmark{7} Center for Quantum Science and Engineering, National Taiwan University, Taipei, Taiwan}
\IEEEauthorblockA{\IEEEauthorrefmark{8} Physics Division, National Center for Theoretical Sciences, Taipei, Taiwan}
\IEEEauthorrefmark{9} \href{mailto:kuoenjui@nycu.edu.tw}{kuoenjui@nycu.edu.tw}.
\IEEEauthorrefmark{10} \href{mailto:goan@phys.ntu.edu.tw}{goan@phys.ntu.edu.tw}.
\vspace{-20pt}
}

\maketitle

\begin{abstract}
Electrocardiogram (ECG) recordings are sensitive biomedical data, limiting the ability of hospitals and wearable devices to share raw signals for centralized model training.
Federated learning addresses this practical privacy constraint by enabling collaborative model training while keeping raw biosignal data at their respective sources. 
However, federated ECG classification remains challenging due to limited client-side samples, imbalanced arrhythmia labels, and non-independent and identically distributed (non-IID) data across clients.
These constraints require classifiers that are both communication-efficient and robust to cross-client distribution shifts.
In this work, we evaluate a hybrid quantum-inspired Kolmogorov--Arnold network (HQKAN) against a multilayer perceptron (MLP) for five-class arrhythmia classification on the MIT-BIH dataset and three-class classification on the INCART dataset under federated averaging (FedAvg).
Across multiple client configurations, HQKAN improves most aggregate and minority-class metrics while using 37.35\% fewer trainable parameters and reducing communication cost by 24.89\% on MIT-BIH; on INCART, it achieves corresponding reductions of 44.81\% and 36.41\%.
These results indicate that HQKAN offers a compact, communication-efficient and robust alternative to the MLP baseline for privacy-aware federated learning on biosignal data.
\end{abstract}

\begin{IEEEkeywords}
federated learning, quantum-inspired Kolmogorov--Arnold networks, electrocardiogram, biosignal processing
\end{IEEEkeywords}

\section{Introduction}
Electrocardiogram (ECG) is widely used for arrhythmia screening\cite{Hannun2019}, but ECG recordings are privacy-sensitive and often distributed across patients, devices, and institutions.
Federated learning (FL) trains a shared model while keeping data local, usually by exchanging model updates rather than raw signals~\cite{McMahan2017FedAvg}.
In this setting, communication, client heterogeneity, and class imbalance become central design constraints~\cite{li2020federated, karimireddy2020scaffold, zhao2018federatednoniid, luo2021no}.

Quantum-inspired models offer a practical means of exploiting the structural features of quantum machine learning (QML)~\cite{biamonte2017quantum,Schuld2019Quantum, Mitarai2018Quantum} without recourse to noisy intermediate-scale quantum (NISQ) devices~\cite{Preskill2018quantumcomputingin}.
Quantum-inspired Kolmogorov--Arnold Networks (QKANs) are compact models that require far fewer trainable parameters than a conventional multilayer perceptron (MLP), using DatA Re-Uploading ActivatioN (DARUAN) edge functions as their learnable activations~\cite{PerezSalinas2020DataReuploading,Schuld2021Effect,jiang2025quantumvariationalactivationfunctions,Sim2019Expressibility}.
The hybrid QKAN (HQKAN), first introduced in~\cite{jiang2025quantumvariationalactivationfunctions}, builds on this by pairing a fully connected encoder and decoder with a QKAN latent feature processor, forming an autoencoder-like architecture.
Since the number of parameters transmitted in each round is a primary driver of communication overhead in FL, this parameter efficiency makes HQKAN particularly well suited to the federated setting~\cite{sidahmed2021efficient, hyeon2021fedpara, konevcny2016federated,jiang2025quantumvariationalactivationfunctions,Wu2022FedKD, Liu2025QT}.

In this work, we investigate whether HQKAN provides a more effective and parameter-efficient classifier than an MLP for federated ECG classification. 
Both models are evaluated under identical client partitions and training budgets across $K\in\{8,16,32\}$ clients on MIT-BIH with the five ANSI/AAMI heartbeat classes and INCART with three ANSI/AAMI classes~\cite{ANSIAAMI_EC57_2012_R2020,incart2008,physionet2000,pollard2026physionet,DeChazal2004Heartbeat}. 
HQKAN uses 11,581 trainable parameters on MIT-BIH and 15,147 on INCART, compared with 18,485 and 27,443, respectively, for the MLP baseline. Across both datasets, HQKAN achieves stronger aggregate and minority-class performance, including higher macro-F1 and Cohen's $\kappa$ and lower Brier scores.

\section{Related Work}

Federated learning trains a shared model across distributed clients while keeping raw data on-device, sharing only model updates and thereby aligning with the privacy requirements of biomedical signals such as ECG, electroencephalogram, and electromyography~\cite{McMahan2017FedAvg,Rieke2020,Sheller2020,Kairouz2021AdvancesFL,Li2020FLChallenges,Zafar2025emg,Baghersalimi2022eeg}.
Applying FL to ECG introduces challenges of non-independent and identically distributed (non-IID) data, class imbalance, and constrained communication, which subsequent studies address through non-IID arrhythmia detection, transfer and semi-supervised learning, and hospital-scale privacy mechanisms~\cite{Zhang2020NonIIDECG,Raza2022FederatedECG,Ying2023FedECG,Baumgartner2023DecentralECG,gutierrez2024applicationfederatedlearningtechniques,Agrawal2024HospitalFLDP,Tang2021PersonalizedECG, Asif2024WeightedFederatedECG, Elmir2025FederatedGAFECG, Islam2025ResourceAwareECG, Sakib2021AsyncECGFL}.
These efforts instantiate the federation with conventional backbones such as convolutional neural networks (CNNs), long short-term memory (LSTM) networks, and residual networks (ResNets)~\cite{healthcare13212811,shu2026bifedkdbidirectionalfederatedknowledge}.

On the other hand, data re-uploading encodes classical inputs into single-qubit rotations repeated across circuit layers~\cite{PerezSalinas2020DataReuploading,Schuld2021Effect}.
QKANs adopt the mechanism of DARUAN with trainable single-qubit circuits as edge functions to replace the fixed spline activations of Kolmogorov--Arnold networks~\cite{jiang2025quantumvariationalactivationfunctions}.
Their applications include solar cycle forecasting, quantum dynamics forecasting, telecommunication forecasting, traffic-matrix forecasting, quantum circuit generation and language modeling~\cite{peng2026qkanfwp,peng2026qkanfwp_TM,hsu2026qkanlstm,lin2026gqkae,jiang2025quantumvariationalactivationfunctions,hsu2024qklstm, hsu2025qaenet, YuCheng2026Meta, Chen_2025_Validating}.

\section{Methods}

\subsection{Dataset and Preprocessing}

The preprocessed MIT-BIH Arrhythmia benchmark~\cite{Moody2001MITBIH} from~\cite{kachuee2018ecgheartbeat} is used as the primary dataset. 
Each heartbeat is represented as a single-lead waveform of length 187, normalized to $[0,1]$, zero-padded when necessary, and categorized into five classes: non-ectopic (N), supraventricular ectopic (S), ventricular ectopic (V), fusion (F), and unknown (Q). 
The original training set is partitioned into 70,043 training and 17,511 validation beats using a seeded label-stratified 80/20 split, while the official 21,892-beat test set is reserved exclusively for final evaluation. 
The dataset is highly imbalanced, with original training-set class counts of 72{,}471 N, 2{,}223 S, 5{,}788 V, 641 F, and 6{,}431 Q beats, corresponding to an approximately 113:1 imbalance ratio between the majority and rarest classes. 
For cross-dataset evaluation, the INCART dataset is processed following the preprocessing protocol of~\cite{ISLAM2024106211}, with the same $[0,1]$ normalization applied to the extracted heartbeat segments. 
The resulting dataset contains 104{,}273 N, 1{,}331 S, and 13{,}586 V beats.

\subsection{Federated Setup}
The federated optimization pipeline is built on FLamby~\cite{NEURIPS2022_232eee8e}, and federated averaging (FedAvg) is used as the only aggregation method.
At round $r \in \{1,\dots,R\}$, client $k$ initializes its local model with the global parameters $\theta^r$, trains for $E$ local epochs, and returns the updated parameters $\theta_k^{r+1}$.
The server aggregates the client models as $\theta^{r+1}=\sum_{k=1}^{K}\frac{n_k}{\sum_{j=1}^{K}n_j}\,\theta_k^{r+1}$, where $n_k$ denotes the number of training beats assigned to client $k$.
Each experimental configuration is defined by a choice of classifier (HQKAN or MLP), client count $K \in \{8, 16, 32\}$, and data partition (IID, or non-IID with concentration $\alpha$). Every configuration uses full client participation in each communication round and is trained under an identical budget of $R=30$ rounds with $E=5$ local epochs.

The two partition regimes are constructed as follows.
In the IID setting, samples from each class are shuffled and distributed to clients in a round-robin manner.
In the non-IID setting, client-specific label proportions are drawn from a Dirichlet distribution with concentration parameter $\alpha$.
The main non-IID experiments are conducted with $\alpha=0.3$, while robustness is assessed by sweeping $\alpha\in\{0.1,0.3,0.5,1,1000\}$.

Each client minimizes a cross-entropy loss weighted by inverse class frequency to counter the severe imbalance across the different AAMI classes, and local updates use AdamW~\cite{loshchilov2017decoupled} with a learning rate of $10^{-3}$, a weight decay of $10^{-4}$, and a batch size of $128$.
A held-out validation set is used solely for model selection: for each configuration we retain the global checkpoint that maximizes the validation macro-averaged area under the precision-recall curve (AUPRC) over the 30 rounds, as it remains sensitive to minority-class performance whereas accuracy is dominated by the majority class.

\subsection{Classifier}
Both models use identical inputs within each dataset and differ only in the classifier architecture. 
For MIT-BIH, the flattened input dimension is 187; the MLP baseline is a fully connected network with ReLU activations and 18{,}485 trainable parameters, whereas HQKAN combines a fully connected encoder and decoder with a QKAN latent feature processor, requiring only 11{,}581 trainable parameters, a 37.35\% reduction. 
For INCART, the corresponding input dimension is 300, with 27{,}443 trainable parameters for the MLP and 15{,}147 for HQKAN, yielding a 44.81\% reduction. 
Since FedAvg synchronizes the full model state, the communication footprint is determined by the total parameter count rather than the trainable parameter count alone; under this measure, HQKAN reduces per-round communication by 24.89\% on MIT-BIH and 36.41\% on INCART. 
HQKAN is implemented using the GPU-efficient \texttt{FlashQKAN} framework in PyTorch~\cite{paszke2019pytorch}, with fused operators and block tiling accelerated through \texttt{cuTe DSL}~\cite{cecka2026cutelayoutrepresentationalgebra}, building on the open-source QKAN implementation~\cite{jiang2025qkan_github}.

\section{Results and Discussion}

Performance is reported with macro-F1, Cohen's $\kappa$, and the Brier score, together with
class-wise positive predictive value (PPV), sensitivity (Sen), and specificity (Spe) under a one-vs-rest convention.
Macro-F1 is our key measure, complemented by the area under the receiver operating characteristic curve (AUROC) and AUPRC.
Client label heterogeneity is summarized by the mean Hellinger distance $H(\mathbf{p},\mathbf{q})
=\tfrac{1}{\sqrt{2}}\big(\sum_{c=1}^{C}(\sqrt{p_c}-\sqrt{q_c})^2\big)^{1/2}$ between each client's
local label distribution $\mathbf{p}$ and the global distribution $\mathbf{q}$, where larger values
($H\in[0,1]$) indicate stronger non-IID skew.

\subsection{Aggregate Performance}
Tables~\ref{tab:agg} and~\ref{tab:agg_incart} report the mean aggregate performance over five seeds on MIT-BIH and INCART, respectively.
Across both datasets, HQKAN consistently outperforms the MLP baseline on all reported aggregate metrics under both IID and non-IID partitions.
For MIT-BIH, the IID macro-F1 advantage of HQKAN increases from 0.013 with 8 clients to 0.054 with 32 clients, suggesting greater robustness as the number of local samples per client decreases.
A similar trend is observed on INCART, where the corresponding gap increases from 0.023 to 0.030.
Under non-IID partitions, HQKAN consistently achieves higher macro-F1 and $\kappa$ and lower Brier scores across all client counts on both datasets.
At the most distributed 32-client setting, HQKAN attains a macro-F1 of 0.761 versus 0.698 for MLP on MIT-BIH, while reducing the Brier score from 0.121 to 0.094.
The same advantage persists on INCART, where HQKAN achieves a macro-F1 of 0.850 compared with 0.838 for MLP and reduces the Brier score from 0.040 to 0.032.
These consistent gains across two ECG benchmarks indicate that HQKAN maintains stronger aggregate predictive performance as federation becomes more fragmented and statistically heterogeneous.

\subsection{Minority-Class Performance}

Tables~\ref{tab:perclass} and~\ref{tab:perclass_incart} report mean class-wise PPV, sensitivity, and specificity at 32 clients. Across both datasets, HQKAN achieved higher mean values on most minority-class measures, although the pattern depended on the class, metric, and partition regime.

On MIT-BIH with IID partitions, the largest mean PPV difference occurred for class F (0.239 vs 0.147). HQKAN also attained higher specificity (0.980 vs 0.962), but lower sensitivity (0.837 vs 0.868). Under non-IID partitioning, HQKAN's mean sensitivity exceeded MLP's by 0.147, 0.077, and 0.073 for classes S, V, and F, respectively. For class S, this difference coincided with slightly lower PPV (0.755 vs 0.768), whereas HQKAN attained higher PPV for V and F.

On INCART, HQKAN's higher mean S-class PPV coincided with lower mean sensitivity under both partitions. Under IID partitioning, its PPV and specificity were higher by 0.079 and 0.013, respectively, while its sensitivity was lower by 0.005. Under non-IID partitioning, its PPV and specificity were higher by 0.083 and 0.001, respectively, while its sensitivity was lower by 0.023. For class V, HQKAN's mean sensitivity was higher by 0.008 under IID partitioning and by 0.020 under non-IID partitioning. Under IID partitioning, its PPV and specificity were lower by 0.021 and 0.004, respectively. Under non-IID partitioning, its PPV was higher by 0.006, while specificity was identical at the reported precision.

\begin{figure*}[t]
    \centering
    \includegraphics[width=\textwidth]{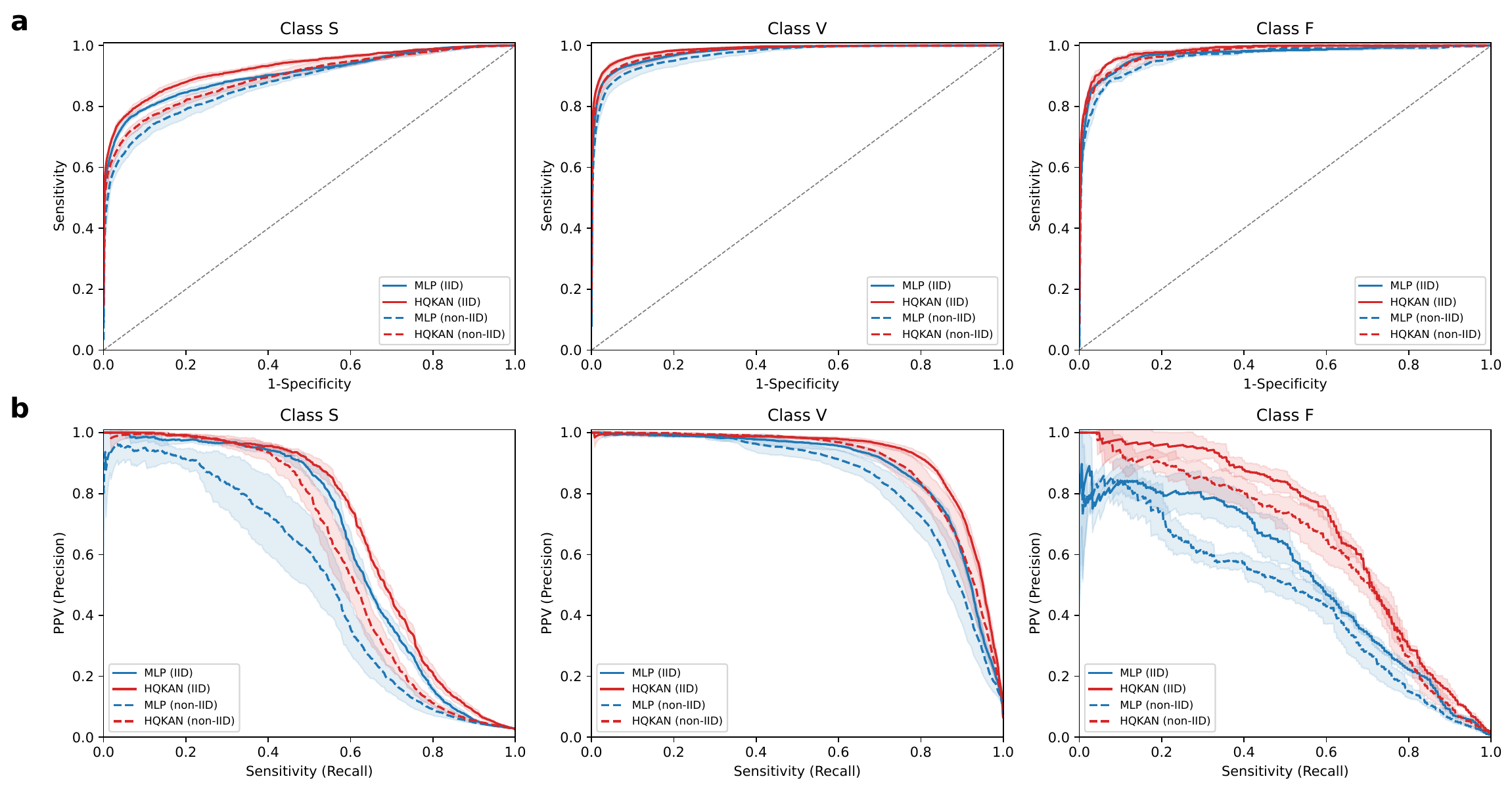}
    \vspace{-10pt}
    \caption{ROC curves (a) and precision-recall curves (b) for the three
    minority classes (S, V, F) under IID and non-IID federated partitions on MIT-BIH,
    comparing the MLP baseline against HQKAN.
    Curves are averaged over five random seeds.}
    \label{fig:roc_prc}
\vspace{-15pt}
\end{figure*}

\begin{figure*}[t]
    \centering
    \includegraphics[width=\textwidth]{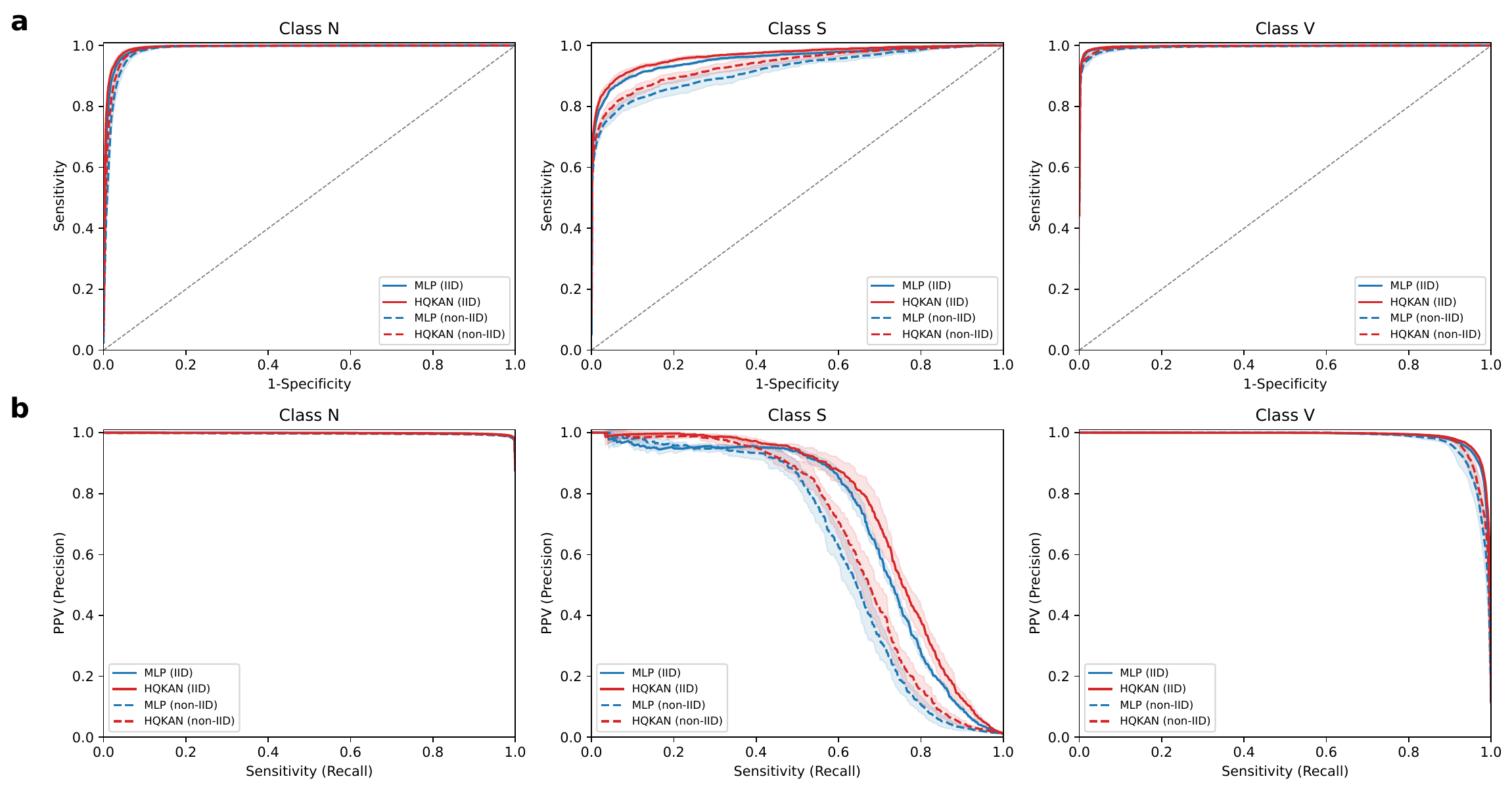}
    \vspace{-10pt}
    \caption{ROC curves (a) and precision-recall curves (b) for the majority class N and minority classes S and V under IID and non-IID federated partitions on INCART, comparing the MLP baseline with HQKAN. Curves are averaged over five random seeds.}
    \label{fig:roc_prc_incart}
\vspace{-15pt}
\end{figure*}

The class-wise precision-recall curves in Figs.~\ref{fig:roc_prc} and~\ref{fig:roc_prc_incart} further characterize these trade-offs across decision thresholds. Visually, model separation was most pronounced for class F on MIT-BIH and class S on INCART.

\subsection{Robustness to Client Heterogeneity}

\begin{figure}[t]
  \centering
  \includegraphics[width=\columnwidth]{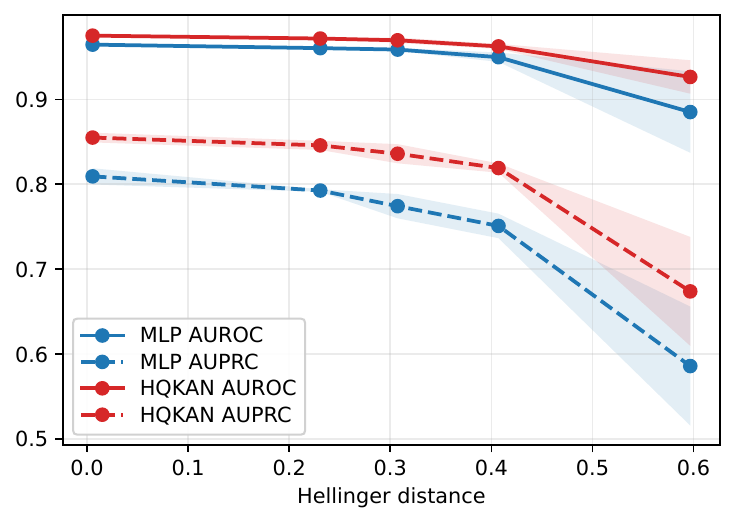}
  \vspace{-12pt}
  \caption{Robustness to client label heterogeneity at 32 clients and 5 local epochs of MIT-BIH dataset.
  The x-axis uses Hellinger distance to summarize label skew, with larger values indicating stronger non-IID partitions.
  The y-axis reports AUROC or AUPRC of different models.
  Each point corresponds to one Dirichlet concentration $\alpha$.}
  \label{fig:robustness_hd}
\vspace{-15pt}
\end{figure}

\begin{figure}[t]
  \centering
  \includegraphics[width=\columnwidth]{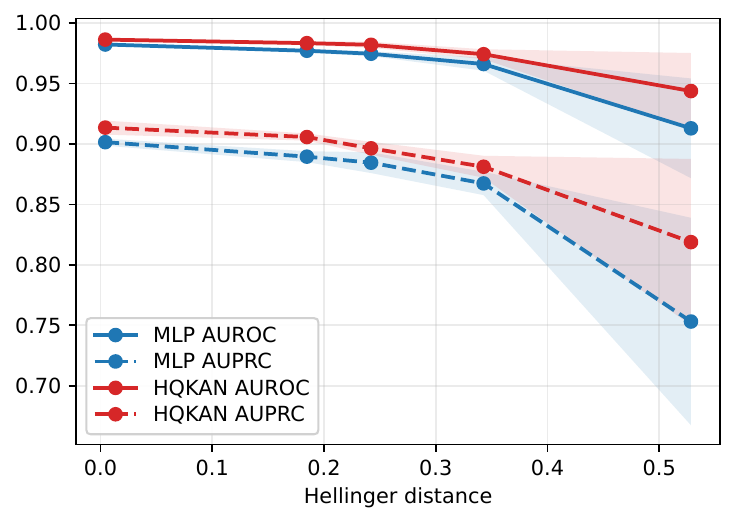}
  \vspace{-12pt}
  \caption{Robustness to client label heterogeneity at 32 clients and 5 local epochs of INCART dataset.
  The x-axis uses Hellinger distance to summarize label skew, with larger values indicating stronger non-IID partitions.
  The y-axis reports AUROC or AUPRC of different models.
  Each point corresponds to one Dirichlet concentration $\alpha$.}
  \label{fig:robustness_hd_incart}
\vspace{-15pt}
\end{figure}

Figures~\ref{fig:robustness_hd} and~\ref{fig:robustness_hd_incart} compare both models at 32 clients across five Dirichlet concentrations. Hellinger distance summarizes heterogeneity across client label distributions, with larger values indicating stronger label skew. At every tested concentration, HQKAN attained higher mean AUROC and AUPRC than MLP on both datasets. The margins widened at the largest Hellinger distances, particularly for AUPRC, where MLP's mean performance declined more sharply. These results support HQKAN's robustness relative to MLP under simulated label-distribution skew.

\begin{table}[t]\centering\footnotesize
\caption{Performance of MIT-BIH dataset with 5 local epochs (mean$\pm$std over 5 seeds). Non-IID uses $\alpha=0.3$.}
\label{tab:agg}
\setlength{\tabcolsep}{3pt}
\begin{tabular}{@{}lllccc@{}}\toprule
Setting & Clients & Model & macro-F1$\uparrow$ & $\kappa\uparrow$ & Brier$\downarrow$ \\
\midrule
\multirow{7}{*}{IID} & \multirow{2}{*}{8} & MLP & 0.726$\pm$0.008 & 0.764$\pm$0.011 & 0.127$\pm$0.007 \\
 &  & HQKAN & \textbf{0.739$\pm$0.006} & \textbf{0.776$\pm$0.008} & \textbf{0.119$\pm$0.005} \\
 \cmidrule{2-6}
 & \multirow{2}{*}{16} & MLP & 0.672$\pm$0.007 & 0.686$\pm$0.013 & 0.178$\pm$0.009 \\
 &  & HQKAN & \textbf{0.716$\pm$0.018} & \textbf{0.744$\pm$0.025} & \textbf{0.137$\pm$0.017} \\
  \cmidrule{2-6}
 & \multirow{2}{*}{32} & MLP & 0.633$\pm$0.008 & 0.626$\pm$0.013 & 0.232$\pm$0.011 \\
 &  & HQKAN & \textbf{0.687$\pm$0.012} & \textbf{0.701$\pm$0.017} & \textbf{0.164$\pm$0.011} \\
\midrule
\multirow{7}{*}{Non-IID} & \multirow{2}{*}{8} & MLP & 0.682$\pm$0.096 & 0.728$\pm$0.103 & 0.130$\pm$0.047 \\
 &  & HQKAN & \textbf{0.718$\pm$0.069} & \textbf{0.761$\pm$0.073} & \textbf{0.112$\pm$0.029} \\
  \cmidrule{2-6}
 & \multirow{2}{*}{16} & MLP & 0.696$\pm$0.059 & 0.771$\pm$0.051 & 0.107$\pm$0.018 \\
 &  & HQKAN & \textbf{0.745$\pm$0.047} & \textbf{0.793$\pm$0.068} & \textbf{0.097$\pm$0.026} \\
  \cmidrule{2-6}
 & \multirow{2}{*}{32} & MLP & 0.698$\pm$0.030 & 0.748$\pm$0.026 & 0.121$\pm$0.010 \\
 &  & HQKAN & \textbf{0.761$\pm$0.026} & \textbf{0.804$\pm$0.027} & \textbf{0.094$\pm$0.012} \\
\bottomrule
\end{tabular}
\vspace{-10pt}
\end{table}

\begin{table}[t]\centering\footnotesize
\caption{Performance of INCART dataset with 5 local epochs (mean$\pm$std over 5 seeds). Non-IID uses $\alpha=0.3$.}
\label{tab:agg_incart}
\setlength{\tabcolsep}{3pt}
\begin{tabular}{@{}lllccc@{}}\toprule
Setting & Clients & Model & macro-F1$\uparrow$ & $\kappa\uparrow$ & Brier$\downarrow$ \\
\midrule
\multirow{7}{*}{IID} & \multirow{2}{*}{8} & MLP & 0.815$\pm$0.015 & 0.882$\pm$0.013 & 0.047$\pm$0.007 \\
 &  & HQKAN & \textbf{0.838$\pm$0.023} & \textbf{0.898$\pm$0.017} & \textbf{0.040$\pm$0.009} \\
 \cmidrule{2-6}
 & \multirow{2}{*}{16} & MLP & 0.764$\pm$0.014 & 0.824$\pm$0.017 & 0.071$\pm$0.007 \\
 &  & HQKAN & \textbf{0.806$\pm$0.036} & \textbf{0.864$\pm$0.036} & \textbf{0.054$\pm$0.011} \\
  \cmidrule{2-6}
 & \multirow{2}{*}{32} & MLP & 0.743$\pm$0.014 & 0.793$\pm$0.026     & 0.090$\pm$0.011 \\
 &  & HQKAN & \textbf{0.773$\pm$0.032} & \textbf{0.827$\pm$0.039} & \textbf{0.068$\pm$0.014} \\
\midrule
\multirow{7}{*}{Non-IID} & \multirow{2}{*}{8} & MLP & 0.788$\pm$0.065 & 0.804$\pm$0.157 & 0.066$\pm$0.043 \\
 &  & HQKAN & \textbf{0.816$\pm$0.061} & \textbf{0.842$\pm$0.101} & \textbf{0.055$\pm$0.035} \\
  \cmidrule{2-6}
 & \multirow{2}{*}{16} & MLP & 0.787$\pm$0.089 & 0.882$\pm$0.030 & 0.044$\pm$0.013 \\
 &  & HQKAN & \textbf{0.835$\pm$0.052} & \textbf{0.893$\pm$0.056} & \textbf{0.037$\pm$0.015} \\
  \cmidrule{2-6}
 & \multirow{2}{*}{32} & MLP & 0.838$\pm$0.009 & 0.891$\pm$0.015 & 0.040$\pm$0.006 \\
 &  & HQKAN & \textbf{0.850$\pm$0.026} & \textbf{0.910$\pm$0.008} & \textbf{0.032$\pm$0.003} \\
\bottomrule
\end{tabular}
\vspace{-10pt}
\end{table}

\begin{table}[t]\centering\footnotesize
\caption{Per-class performance at 32 clients and 5 local epochs of MIT-BIH dataset (mean$\pm$std over 5 seeds). Non-IID uses $\alpha=0.3$.}
\label{tab:perclass}
\setlength{\tabcolsep}{3pt}
\begin{tabular}{@{}lllccc@{}}\toprule
Setting & Class & Model & PPV & Sen & Spe \\
\midrule
\multirow{7}{*}{IID} & \multirow{2}{*}{S} & MLP & 0.282$\pm$0.009 & 0.739$\pm$0.018 & 0.951$\pm$0.003 \\
&  & HQKAN & \textbf{0.334$\pm$0.030} & \textbf{0.747$\pm$0.014} & \textbf{0.961$\pm$0.006} \\
\cmidrule{2-6}
& \multirow{2}{*}{V} & MLP & 0.669$\pm$0.017 & 0.881$\pm$0.003 & 0.969$\pm$0.002 \\
&  & HQKAN & \textbf{0.705$\pm$0.019} & \textbf{0.908$\pm$0.012} & \textbf{0.973$\pm$0.003} \\
\cmidrule{2-6}
& \multirow{2}{*}{F} & MLP & 0.147$\pm$0.015 & \textbf{0.868$\pm$0.007} & 0.962$\pm$0.004 \\
&  & HQKAN & \textbf{0.239$\pm$0.022} & 0.837$\pm$0.014 & \textbf{0.980$\pm$0.003} \\
\midrule
\multirow{7}{*}{Non-IID} & \multirow{2}{*}{S} & MLP & \textbf{0.768$\pm$0.114} & 0.368$\pm$0.138 & \textbf{0.996$\pm$0.004} \\
&  & HQKAN & 0.755$\pm$0.089 & \textbf{0.515$\pm$0.054} & 0.995$\pm$0.003 \\
\cmidrule{2-6}
& \multirow{2}{*}{V} & MLP & 0.827$\pm$0.053 & 0.721$\pm$0.075 & 0.989$\pm$0.005 \\
&  & HQKAN & \textbf{0.838$\pm$0.045} & \textbf{0.798$\pm$0.034} & 0.989$\pm$0.004 \\
\cmidrule{2-6}
& \multirow{2}{*}{F} & MLP & 0.296$\pm$0.071 & 0.680$\pm$0.088 & 0.987$\pm$0.006 \\
&  & HQKAN & \textbf{0.380$\pm$0.113} & \textbf{0.753$\pm$0.065} & \textbf{0.989$\pm$0.007} \\
\bottomrule
\end{tabular}
\vspace{-10pt}
\end{table}

\begin{table}[t]\centering\footnotesize
\caption{Per-class performance at 32 clients and 5 local epochs of INCART dataset (mean$\pm$std over 5 seeds). Non-IID uses $\alpha=0.3$.}
\label{tab:perclass_incart}
\setlength{\tabcolsep}{3pt}
\begin{tabular}{@{}lllccc@{}}\toprule
Setting & Class & Model & PPV & Sen & Spe \\
\midrule
\multirow{7}{*}{IID} & \multirow{2}{*}{N} & MLP & 0.994$\pm$0.000 & 0.945$\pm$0.010 & 0.961$\pm$0.003 \\
&  & HQKAN & \textbf{0.995$\pm$0.001} & \textbf{0.956$\pm$0.013} & \textbf{0.966$\pm$0.004} \\
\cmidrule{2-6}
& \multirow{2}{*}{S} & MLP & 0.199$\pm$0.025 & \textbf{0.838$\pm$0.009} & 0.961$\pm$0.008 \\
&  & HQKAN & \textbf{0.278$\pm$0.076} & 0.833$\pm$0.013 & \textbf{0.974$\pm$0.009} \\
\cmidrule{2-6}
& \multirow{2}{*}{V} & MLP & \textbf{0.916$\pm$0.014} & 0.966$\pm$0.004 & \textbf{0.989$\pm$0.002} \\
&  & HQKAN & 0.895$\pm$0.038 & \textbf{0.974$\pm$0.005} & 0.985$\pm$0.006 \\
\midrule
\multirow{7}{*}{Non-IID} & \multirow{2}{*}{N} & MLP & 0.983$\pm$0.004 & 0.991$\pm$0.004 & 0.877$\pm$0.029 \\
&  & HQKAN & \textbf{0.985$\pm$0.004} & \textbf{0.994$\pm$0.002} & \textbf{0.893$\pm$0.027} \\
\cmidrule{2-6}
& \multirow{2}{*}{S} & MLP & 0.725$\pm$0.183 & \textbf{0.545$\pm$0.105} & 0.997$\pm$0.004 \\
&  & HQKAN & \textbf{0.808$\pm$0.105} & 0.522$\pm$0.120 & \textbf{0.998$\pm$0.001} \\
\cmidrule{2-6}
& \multirow{2}{*}{V} & MLP & 0.956$\pm$0.014 & 0.905$\pm$0.029 & 0.995$\pm$0.002 \\
&  & HQKAN & \textbf{0.962$\pm$0.015} & \textbf{0.925$\pm$0.022} & 0.995$\pm$0.002 \\
\bottomrule
\end{tabular}
\vspace{-10pt}
\end{table}

\section{Conclusion}
We studied HQKAN as a parameter-efficient classifier for federated ECG classification, benchmarking it against an MLP baseline under matched FedAvg training budgets with 8, 16, and 32 clients across IID and label-skewed partitions of the five-class MIT-BIH and three-class INCART datasets.
Across all evaluated configurations, HQKAN outperforms the baseline in mean macro-F1 and Cohen's $\kappa$ while achieving lower mean Brier scores. 
It also maintains higher AUROC and AUPRC across all evaluated levels of label heterogeneity. 
These gains come with substantially lower model and communication overhead. Against the MLP baseline, HQKAN reduces trainable parameters by 37.35\% and communication cost by 24.89\% on MIT-BIH; on INCART, the corresponding reductions are 44.81\% and 36.41\%.

These results indicate that robustness to client heterogeneity need not come at the expense of model or communication efficiency. Across the evaluated IID and label-skewed federated settings, HQKAN maintains stronger predictive performance while using fewer trainable parameters and incurring lower communication cost than the MLP baseline.

\section*{Acknowledgment}
C.-H. Lin, K.-C. Peng, J.-C. Jiang and Y.-C. Hsu thank the National Center for High-Performance Computing (NCHC), National Institutes of Applied Research (NIAR), Taiwan, for providing computational and storage resources supported by the National Science and Technology Council (NSTC), Taiwan, under Grants No. NSTC 114-2119-M-007-013 and NSTC 115-2119-M-007-005.
H.-S. Goan acknowledges support from the NSTC, Taiwan, under Grants No. NSTC 113-2112-M-002-022-MY3, No. NSTC 113-2119-M-002-021, No. NSTC 114-2119-M-002-018, No. NSTC 114-2119-M-002-017-MY3, and from the National Taiwan University under Grants No. NTU-CC-115L8937, No. NTU-CC-115L893704 and No. NTU-CC-115L8512.
H.-S. Goan is also grateful for the support of the “Center for Advanced Computing and Imaging in Biomedicine (NTU-115L900702)” through the Featured Areas Research Center Program within the framework of the Higher Education Sprout Project by the Ministry of Education (MOE), Taiwan, the support of Taiwan Semiconductor Research Institute (TSRI) through the Joint Development Project (JDP) and the support from the Physics Division, National Center for Theoretical Sciences, Taiwan.
E.-J. Kuo acknowledges financial support from the NSTC of Taiwan under Grant No.~NSTC~114-2112-M-A49-036-MY3.

\bibliographystyle{IEEEtran}
\bibliography{reference}

\end{document}